# Frequency based Classification of Activities using Accelerometer Data

Annapurna Sharma, Amit Purwar, Young-Dong Lee, Young-Sook Lee and Wan-Young Chung

*Abstract* — This work presents, the classification of user activities such as Rest, Walk and Run, on the basis of frequency component present in the acceleration data in a wireless sensor network environment. As the frequencies of the above mentioned activities differ slightly for different person, so it gives a more accurate result. The algorithm uses just one parameter i.e. the frequency of the body acceleration data of the three axes for classifying the activities in a set of data. The algorithm includes a normalization step and hence there is no need to set a different value of threshold value for magnitude for different test person. The classification is automatic and done on a block by block basis.

## I. INTRODUCTION

With the advancement of technology, The traditional health care systems need to be modified so as to give flexibility to a user's lifestyle. For aging population who are living independently, their activity & behavior monitoring can give significant information about their health. An appropriate level of physical activity assists people to develop healthy cardiovascular system, healthy musculoskeletal tissues and maintain a healthy body weight [1], in lowering blood pressure, in increasing the level of the good high-density lipoprotein (HDL) cholesterol, in improving sugar tolerance, and in changing hormone levels to a direction more suitable for preventing cancer [2]. WHO recommends at least 60 minutes of moderate- to vigorous-intensity physical activity to ensure healthy development of school-aged- youth [1].

The series of studies that categorized activities of daily living basically include either vision based or accelerometer based systems. The vision based systems use visual aids to get the posture and movement information. But these systems are expensive and inherently limit the user's mobility to a predefined area. However, the accelerometer based systems that use wireless sensor networks can impart more mobility to the user and are cheaper than the former.

Juha Parakka et al. [3] described methods for classification of activities like cycling, walking, running and rowing. The main aim of this work was to study activity classification, which are the most information-rich sensors and what kind of signal processing and classification methods should be used for activity classification. The study suggested several time and frequency domain features, of the accelerometer data, along with magnetometer data for the classification.

N. Bidargaddi et al. [4] developed a wavelet based algorithm for detecting and calculating the duration of sit-to-stand and stand-to-sit transition from the Signal magnitude vector of the accelerometer data.

D.M. Karantonis et al. [5] presented an algorithm to classify the activities such as rest, movement, fall and postures of the person (Standing, Lying sub postures, and sitting) using accelerometer sensor. The system proposed a majority of signal processing on the wearable unit. The classification was done using the separation of bodily and gravity acceleration components. The system also proposed an indirect measure of metabolic energy expenditure.

A more recent work by James F. Knight et al. [6] showed the use of accelerometers in wearable systems for a number of applications. The work was restricted to the consideration as to how raw data and Time domain features can be used for context awareness.

This paper presents an algorithm for classification of Rest, walk and run activities using just one parameter i.e. Frequency of the received data by an accelerometer system. The algorithm uses normalization in frequency domain and hence there is no need of setting a threshold value of magnitude. Also the system uses Wireless sensor node which makes it suitable for Ubiquitous health care.

## II. SYSTEM DESIGN

### A. System Hardware

The Sensor unit consists of a capacitive type, 3-axis Micro electromechanical sensor (MEMS) accelerometer MMA7260Q (Freescale Inc., USA) along with the Telos type sensor node TIP710 (Maxfor Co. Ltd, Korea). The three axis accelerometer is having an acceleration range of -6g to + 6g and sensitivity of 200mV/g. ('g' is the acceleration due to gravity in m/s$^2$). The sensor node consists of MSP430F1611 microcontroller which has inbuilt 48Kbytes program memory

Manuscript received March 1, 2008. This research was financially supported by the Ministry of Commerce, Industry and Energy(MOCIE) and Korea Industrial Technology Foundation(KOTEF) through the Human Resource Training Project for Regional Innovation.

Annapurna Sharma is with Dept. of Ubiquitous IT, Graduate School of Design & IT, Dongseo University, Busan 617-716, South Korea (e-mail: sharmaannapurna@gmail.com).

Amit Purwar is with Dept. of Ubiquitous IT, Graduate School of Design & IT, Dongseo University, Busan 617-716, South Korea (e-mail: purwaramit12@yahoo.com).

Young-Dong Lee is with Dept. of Ubiquitous IT, Graduate School of Design & IT, Dongseo University, Busan 617-716, South Korea (e-mail: ydlee2@gmail.com).

Young-Sook Lee is with Dept. of Ubiquitous IT, Graduate School of Design & IT, Dongseo University, Busan 617-716, South Korea (e-mail: ulysseslee@dongseo.ac.kr).

Wan-Young Chung is with Division of Computer Information Engineering, Dongseo University, Busan 617-716, South Korea (phone: +82-51-320-1756; fax: +82-51-327-8955; e-mail: wychung@dongseo.ac.kr).

and 10Kbytes RAM. The sensor node uses CC2420 2.4 GHz ISM band for radio communication. It can transfer the data at a rate of 250 kbps per channel with a radio range of 20 to 75 meters depending on the obstacles and on the environment, whether used in indoor or outdoor. The whole sensor unit is powered by two AA batteries.

The base station consists of another similar Telos type node for collecting the packets. This sends the received data to PC by RS-232. Further data analysis and processing is done using MATLAB 7.4.0.

The sampling rate for the sensor unit was chosen to be 50 Hz so as to fulfill the Nyquist criterion [7], keeping an eye to the fact that most of the motion artifacts lie within the frequency range of 20 Hz [5]. Most of the signal strength for the accelerometer data was found to lie within 0-15 Hz and maximum high amplitude peaks of the considered activities were found up to a range of 5 Hz as shown in figure 1.

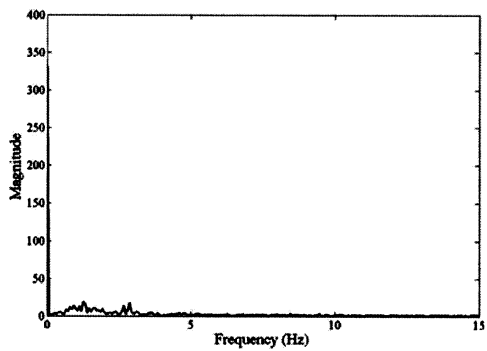

Fig. 1. Frequency response of Z-axis data during running.

### B. Software

The software for sensor unit and Base station sensor is developed using nesC as the programming language, with Tiny-OS [8] as the real time operating system. Tiny-OS handles various resources like memory and processing power and keeps synchronization between various events. Tiny-OS is a component based, event driven operating system in which tasks perform primary work and these tasks are atomic, i.e. tasks can not be preempted by one another, they run up to completion, with respect to each other but can be interrupted by events. Events are the hardware interrupts. Priority is given to the hardware interrupt like interrupt from ADC or radio over the data processing task. At lower level components have handlers connected directly to hardware interrupt which can be external interrupts, timer events or counter events.

## III. METHODS

### A. Calibration

The accelerometer data received from the tri-axial accelerometer sensor are sampled, quantized and packetized for radio transmission. At the base station PC, these quantized data is needed to be converted back to acceleration values before any further processing. A linear calibration is used to convert the data values to acceleration in terms of 'g', the acceleration due to gravity in m/s$^2$. The device is so calibrated as to give a value of acceleration equal to 0g for both the x and y axis, when the device is positioned vertically. And +1g and -1g for the z-axis when the device is placed vertical upright and vertical inverted respectively. For calibration, the algorithm uses initial 100 samples of the three axis data. Figure 2(a) showing the calibrated data of the three axes.

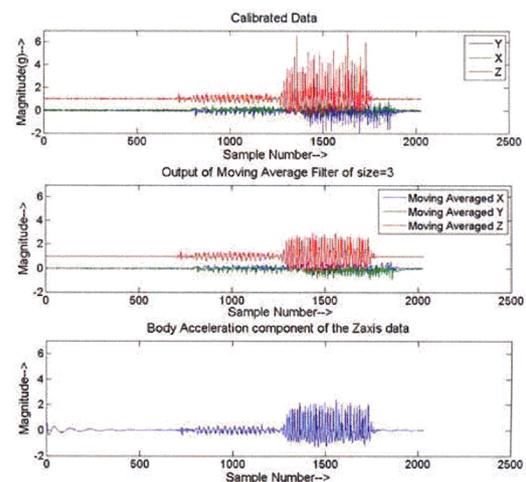

Fig. 2. Calibrated (a) and preprocessed (b) data, (c) showing body acceleration component of the Z-axis data.

### B. Preprocessing

The preprocessing of data is required to make the data appropriate for the classification. We used a two step preprocessing. The first step is to make the raw calibrated data smoother by passing it through a moving average filter. The filter of order 3 is used because it smoothens the acceleration data but does not deform the amplitudes of accelerations so much (Figure 2(b)).

The second step of the preprocessing is to separate out the acceleration signal which is related to the human activity from the received acceleration data. The data received at the base station consists of acceleration due to gravity, due to body acceleration and spurious external noise. So, for getting the information regarding the activity being performed by user, the body acceleration signal needs to be separated out. The frequency of gravity acceleration is found to be limited within the frequency range 0- 0.8 Hz. So a high pass, elliptic filter of order 7 with cutoff frequency=0.5Hz is used to get the body acceleration components from the moving averaged data. Figure 2(c) is showing the body acceleration component of the Z-axis data. The choice of filter is made after testing a number of activity data. For further processing, the body acceleration component of the three axis are combined to give the RMS Body acceleration component (BA) as given by equation (1)

$$RMS = \sqrt{X_B^2 + Y_B^2 + Z_B^2} \qquad (1)$$

Where $X_B$, $Y_B$ and $Z_B$- are the corresponding body acceleration components.

*C. Parameters for Classification*

As shown in figure 2(a), the three axis data consisting of the rest, walk and run activity sequence have different amplitudes of acceleration. The magnitude of acceleration varies with the pace of the activity and is different for different users. Also Figure 3 shows that the frequency of the rest lies within 0.5 Hz, walk in the range 1.5-2.5Hz and run in the range 2.5-4Hz. And these frequency ranges does not vary much with user whose activities are to be monitored. So the algorithm uses frequency as the only parameter for classification.

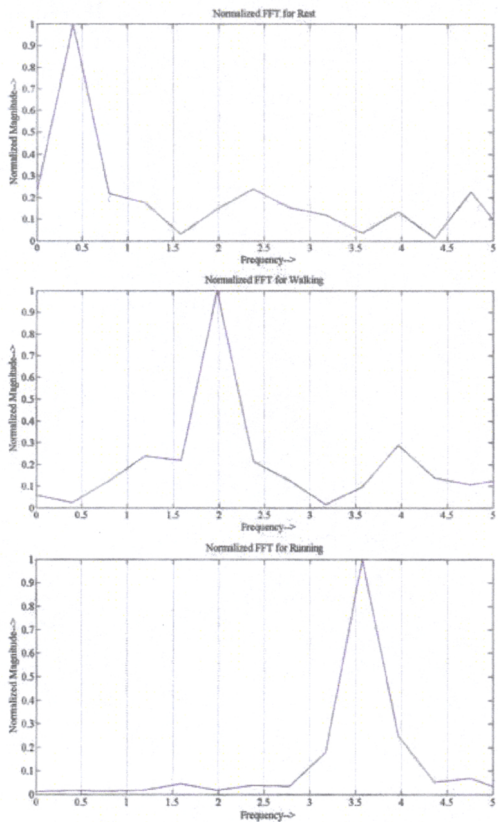

Fig. 3. Normalized FFT for rest, walk and Run, showing the peak in different frequency ranges.

*D. Classification Algorithm*

The classification algorithm receives the three axis accelerometer data from a text file. The data is then calibrated and preprocessed (described in section III.A and III.B) for further classification. The classification is done on a step by step basis. The entire data set is divided into blocks, which then processed to find out the activity artifacts. Each data is processed one at a time. The size of data block is chosen as 64, (approximately 1.28s data) which is an optimal size for classification of activities [6]. The block size is deliberately chosen a power of 2, so that an FFT algorithm can be implemented [7].

The FFT for the entire block is calculated and normalized so as to limit the magnitude axis in the range 0-1. The normalization is done so as to detect a peak without setting a threshold, which implies user independency. The Normalized FFT is then checked for the range of frequency at the peak of the magnitude. The desired ranges for Rest, walk and run have been described in section III.C. In the algorithm, a higher priority, to check for a peak, is given to Run over others and then to Walk and Rest respectively. If the normalized FFT does not show a peak in the desired frequency ranges, it is classified as miscellaneous, which may represent a corrupted data or some other activity which is not under consideration as of now. After declaring the activity in present block, the algorithm moves to the next block in sequence for classification. The algorithm is developed using MATLAB7.4.0. The flow chart for the algorithm is shown in figure4.

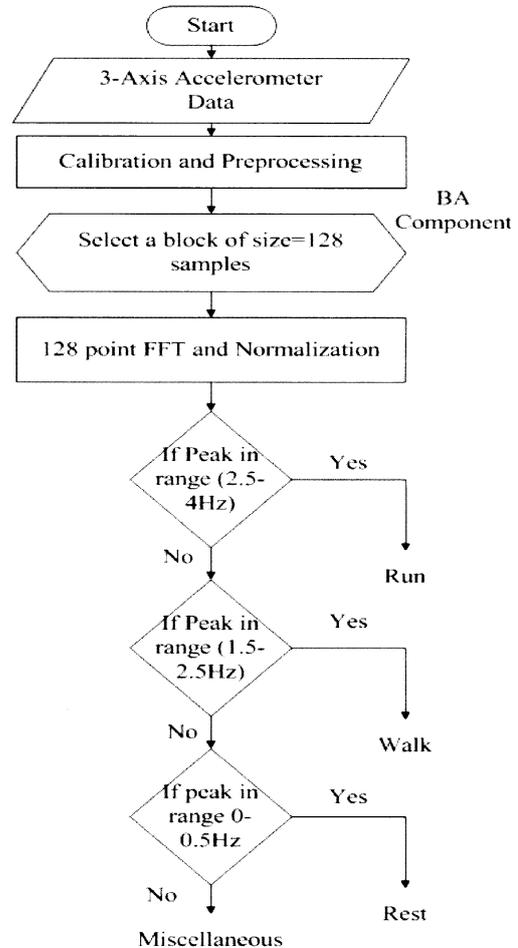

Fig. 4. Flow chart of the classification algorithm.

## IV. EXPERIMENTAL SET UP AND RESULTS

The tests were conducted for two test persons wearing the sensor unit on chest. The data are collected for the activity sequence REST-WALK-RUN in the same order only.

The number of blocks in each data set is different because the number of samples collected for the activities are different for different data set. The classification accuracy is calculated on the basis of comparison with manual retrogression of data and the algorithm output, on a block by block basis. The reason of misclassification of certain blocks is seen to have more than one activities present in that block. The performance results are shown in Table1.

TABLE I
PERFORMANCE TABLE FOR THE CLASSIFICATION ALGORITHM SHOWN IN FIGURE 4

| Test Person--> | A | B | A+B |
|---|---|---|---|
| No. of data sets | 7 | 10 | 17 |
| Total no. of blocks in the data sets | 36 | 137 | 173 |
| No of Blocks classified correctly | 34 | 127 | 161 |
| No. of Blocks classified wrongly | 2 | 10 | 12 |
| Accuracy % | 94.4% | 92.7% | 93.1% |

## V. CONCLUSION

Activity classification is implemented with tri-axial accelerometer and wireless sensor node. Accuracy of the overall classification was detected 93.1%. Detection accuracy shows the feasibility of the system for practical implementation. Some body postures of standing and sitting and other fatal activities like fall can also be included, with the combination of other feature parameters, to extend the application. System is compatible for ubiquitous healthcare using sensor network due to the use of sensor node. Although the system is tested using single hop communication, but can further be improved for multi-hop environment.

REFERENCES

[1] http://www.who.int/dietphysicalactivity/factsheet_young_people/en/
[2] F. Cardenas, R. K. Pon and Robert B. Cameron, "Management of Streaming Body Sensor Data for Medical Information Systems", in Proc. METMBS 2003, Las Vegas, NV, pp. 186- 91.
[3] Juha Parkka, Miikka Ermes, Panu Korpipaa, Jani Mantyjarvi, Johannes Peltola and Ilkka Korhonen, "Activity classification using Realistic data from wearable sensors", IEEE Transactions on Information Technology in Biomedicine 2006, vol.10 no.1, pp 119-128.
[4] Niranjan Bidargaddi, Lasse Klingbeil, Antti Sarela, Justin Boyle, Vivian Cheung, Catherine Yelland, Mohanraj Karunanithi and Len Gray, "Wavelet based approach for posture transition estimation using a waist worn accelerometer", 29th IEEE EMBS Annual International Conference 2007.
[5] D.M. Karantonis, M. R. Narayanan, M. Mathie, N. H. Lovell, and Branko G. Celler, "Implementation of a Real-Time Human Movement Classifier Usinga Triaxial Accelerometer for Ambulatory Monitoring", IEEE Trasactions on Information Technology in Biomedicine, vol.10, no. 1, January 2006.
[6] James F. Knight, Huw W. Bristow, Stamatina Anastopoulou, Chris Baber, Anthony Schwirtz, Theodoros N. Arvanitis, "Uses of accelerometer data collected froma wearable system", pres Ubiquit comput(2007) 11:117-132.
[7] Alan V. Oppenheim, R.W. Schafer, and John R. Buck, Discrete-Time Signal Processing, 2nd Ed., Prentice Hall (1998), p.140.
[8] http://www.tinyos.net/
[9] Wan-Young Chung, Sachin Bhardwaj, Amit Purwar and Dae-Seok Lee, Member, IEEE , "A Fusion Health Monitoring Using ECG and Accelerometer sensors for Elderly Persons at Home", Engineering in Medicine and Biology Society, 29th Annual International Conference of the IEEE, 2007. EMBS 2007.